\documentclass{article}
\usepackage{amsfonts}
\usepackage{graphicx}
\newcommand{\R}{\mathbb{R}}

\begin{document}

\title{Tag Clouds for Displaying Semantics: \\
The Case of Filmscripts}

\author{F. Murtagh, Department of Computer Science, Royal Holloway, 
University
 of London \\ 
Egham TW20 0EX, UK  ({\tt fmurtagh@acm.org}) and Science Foundation Ireland, \\
Wilton Park House, Wilton Place, Dublin 2, 
Ireland \\
 \\
A. Ganz, Department of Media Arts, Royal Holloway, University of London, \\
Egham TW20 0EX, UK  ({\tt adam.ganz@rhul.ac.uk})  \\
\\
S. McKie, Department of Media Arts, Royal Holloway, University of London, \\
Egham TW20 0EX, UK  ({\tt sm@tripos.biz}) \\ 
 \\
J. Mothe, Institut de Recherche en Informatique de Toulouse, \\
118, Route de Narbonne, 31062 Toulouse Cedex 04, France
({\tt mothe@irit.fr}) \\
 \\
K. Englmeier, Fachhochschule Schmalkalden, Blechhammer, \\
98574 Schmalkalden, Germany ({\tt k.englmeier@fh-sm.de})  }

\maketitle

Type of submission: paper.

\bigskip

Total number of pages: 23.

\bigskip

Corresponding author: F. Murtagh.

\subsection*{Acknowledgements}

This work was started in the EU Sixth
Framework project, ``WS-Talk, Web services communicating in the language 
of their community'', 2004--2006.  Pedro Contreras and Dimitri Zervas
contributed to this work.  We thank the anonymous referees for 
valuable comments on an earlier version of this paper.  

\bigskip

\noindent
{\bf Statement of exclusive submission:} 
This paper (or a similar version) is not 
currently under review by a journal or conference, nor will it be submitted
to such within the next three months. 

\newpage

\begin{center}
\section*{Tag Clouds for Displaying Semantics: \\
The Case of Filmscripts}
\end{center}

\begin{abstract}
We relate tag clouds to other forms of visualization, including 
planar or reduced dimensionality  mapping, and Kohonen self-organizing maps.
Using a modified tag cloud visualization,
we incorporate other information into it, including text sequence and 
most pertinent words.  Our notion of word pertinence goes beyond just
word frequency and instead takes a word in a mathematical sense as 
located at the 
average of all of its pairwise relationships.  We capture semantics 
through context, taken as all pairwise relationships.  
Our domain of application is
that
of filmscript analysis.  The analysis of filmscripts, always important for
cinema, is experiencing a major gain in importance in the context of 
television.  
Our objective in this work is to visualize the semantics
of filmscript, and beyond filmscript any other partially structured,
 time-ordered, sequence of text segments.  In particular we develop 
an innovative approach to plot characterization.  

\end{abstract}

\bigskip

\noindent
{\bf Keywords:} 
Correspondence Analysis, semantics, context,  
text analysis, information display, tag cloud, visualization, filmscript


\section{Introduction}
\label{sect1}

\subsection{Visual, Interactive User Interfaces for Supporting Information
Space Navigation}

Visualization is often an important way to elucidate semantic heterogeneity 
for the user.  Visual user interfaces are 
discussed in Murtagh et al.\ (2003), with examples that include 
interactive, responsive information maps based on the Kohonen self-organizing
feature map, and semantic network graphs.  A study is presented in Murtagh 
et al.\ (2003) of such maps used for 
client-side visualization of concept hierarchies relating
to an  economics information space.  

The use of ``semantic road maps'' 
to support information retrieval goes back to Doyle (1961).   Motivation,
following Murtagh et al.\ (2003), includes the following: 
(i) Visualization of the semantic
structure and content of the data store allows the user to have some idea
before submitting a query as to what type of outcome is possible.  Hence
visualization is used to summarize the contents of the database or data
collection (i.e., information space).  (ii) The user's information 
requirements are often fuzzily and ambiguously 
defined at the outset of the information 
search.  Hence visualization is used to help the user in his/her information
navigation, by signaling the relationships between concepts.  (iii) 
Visualization therefore helps the user before the user interacts 
with the information space, and during this interaction.  It is a natural 
enough progression that the visualization can become the user interface.  

Olston and Chi (2003) investigate how the visual user interface design 
can be influenced by the ``semantic road map'' that is being followed by
the user. Olston and Chi (2003) deal with browsing and/or 
search, based on unstructured text and hypertext. Instead, in this 
work, we deal with semi-structured textual data -- filmscripts.  
Other than the immediate benefits relating to the early stages of 
movie or television program creation, there is great potention in our 
work for all areas that use semi-structured textual segments (e.g.\
in medical doctor-patient interactions, or for business or other 
interviews).

Since the mid-1990s we built visual interactive maps of bibliographic and
database information at Strasbourg Astronomical Observatory, and some of
these, with references, are available at Murtagh (2006).  
A comprehensive  view of the Kohonen self-organizing feature map used  
 can be found at websom.hut.fi (Kohonen et al.\ 2000).

A more recent development has been tag clouds.  
McKie (2007) discusses examples and provides an online system for 
creating of filmscript term ``clouds''.  He discusses similar 
tools (e.g., TagCrowd, www.tagcrowd.com; Zoomcloud, zoomclouds.egrupos.net).
Similar tag clouds are commonly used 
to present information in large data repositories (e.g.\ flickr, 
www.flickr.org).    

The motivation for such tools is to have (possibly interactive) annotated maps 
to support information navigation.  Prominent terms are graphically presented 
and can be used to carry out a local search.  In some cases, the location of 
terms is important, in particular in the case of the Kohonen map.  
The {\em automated} annotation of such information maps is not easy.  Often 
the basis for display font size and sometimes even for location on the maps 
is simply frequency of term occurrence.  The work presented in this article
aims at taking more available information into account,
leveraging interrelationships in textual content and thereby semantic 
content.  

Our concern is not with overly large visualizations (see Kaser and 
Lemire, 2007, who use an optimal display layout algorithm) but rather
with (i) structure in the form of sequence, and (ii) taking context and 
thereby semantics into account.  

\subsection{Filmscript}

A filmscript, expressing a story, is the starting point for any possible 
production for cinema or TV.  TV episodes in the same series may each be 
developed by different scriptwriters, and later by different producers 
and directors. The aim of any TV screenplay is to provide a unique but 
repeatable experience in which each episode shares certain structural 
and narrative traits with other episodes from the same series despite 
the fact they may have originated or been realised by different people 
or teams. There is a productive tension between the separate needs for 
uniqueness, -- that each episode seems fresh and surprising, and 
belonging to its genre. An episode of any series needs to 
have a common feel, to offer the specific kinds of pleasure the audience 
associates with the series.  We believe that these distinctive qualities 
of any individual script and the distinctive qualities of any genre could 
be subject to analysis through a tool which finds distinctive ways of 
representing the essential structural qualities of any script, and the 
series to which it belongs and thus enables the writer, the script 
developer or producer to have a deeper understanding of the script 
and have objective criteria for the creative decisions they take. 
Moreover as the scripts are migrated to digital formats the tools 
offer many possibilities for prototyping from the information gathered.
By analysis of multiple screenplays, TV episodes and genres the 
technology will allow the possibility of creating distinctive 
analytical patterns for the structure of genres, series, or 
episodes in the same way that comparative authorship can be 
assessed for individual writers. 

There are major changes taking place in the television and film 
production process.  This profound revolution is taking place in 
both the film and television production areas, and it has significant 
knock-on implications for new digital media areas such as the games 
sector, digital learning environments, and virtual communities or 
societies (e.g.\ Second Life).  Convergence means that many media 
products move between media, a show such as the Simpsons or Lost 
is originated as a TV show and migrated to other media (film, game, 
online). Shows are created in one medium and are spun off to others 
-- from game to film or TV.  Even more symptomatic of changes at 
stake is that the young YouTube and FaceBook generation is well 
attuned to interactive and realtime entertainment. 

\subsection{Free Text, Semi-Structured Text and Filmscripts}
\label{sect12}

Before one considers hyperlinks, text is naturally viewed as sequential.
There may well be other forms of structure beyond hypertext and 
classical sequential text and we will touch
on some in this section -- hierarchical structure in text, 
for example.  The sequential 
text that we target in this work is the filmscript that forms the basis
and starting point 
of a film.  There are various aspects of structure that we consider.  Our
interest lies therefore in semi-structured data visualization and analysis.  

Beard et al.\ (2008) consider data that is both (i) spatially referenced, 
and (ii) temporally referenced, i.e.\ spatio-temporal data analysis.  In 
our work we consider text segments that are characterized by 
(i) free text index terms; and (ii) the text segments are in sequence.  

Let us review some quite general aspects of the handling of sequential 
segments of text.  We then proceed to the particular case of filmscripts.  

Chafe (1979) considers linear versus hierarchical (e.g., at sentence, 
paragraph, section, etc.\ levels) organization of text, in the context
of studying narrative in its role as expressing past experience.  
Chafe used
a 7-minute 16 mm color movie, with sound but no speech, and collected
narrative reminiscences of it from human subjects.   Chafe argues in 
favor of a ``flow model'', i.e.\ a 
``flow of thought and the flow of language'' which ignores all structure
beyond the sequential order. 

In our work, based on film or television movie scripts, we have and avail 
of given scene boundaries.  For Chafe, and others basing their work on a 
similar principle such as Hearst (1994), 
this was not the case and instead they based their work on the 
human thinking that lay behind the recorded narrative.  It will be 
informative and very useful for us to avail instead of the structure 
that is provided by a film or television program script.  

There are literally thousands of film scripts, including for television 
programs, for all genres, 
available and openly accessible on the web (e.g.\ IMSDb, Internet Movie 
Script Database).  A film script is composed of a succession of scenes,
each of which has a header (often in upper case, and indented) 
indicating internal or external, day or night,
location and other metadata, together with transition (``cut to:'', 
``sound cue'') and beginning and end details.   
The variable length scene itself
contains dialog between characters, and/or action description.  

Supporting movie script analysis, both individually and comparatively,  
is of major importance in the distributed and collaborative 
writing process.  Indeed machine learning algorithms 
have been directly applied to scripts themselves to predict later commercial 
success (Gladwell 2006).  

Filmscripts are comprised of semi-structured data, 
incorporating free text and additional structure. 
They provide a very good ``model'' or exemplar for other application fields.  
Analysis, retrieval and use of scripts could provide a model 
for medical report handling, and scenario analysis in 
organization and management.  The scripts may provide a more malleable 
basis for reshaping and restructuring content in order to support 
interactive training and learning environments, as well as the full gamut of
interactive media in entertainment.  

An in-depth analysis of content of 
the film Casablanca can be found in Murtagh et al.\ (2009, and see 
discussion in Merali, 2008).  Television 
presents new and interesting challenges for a whole range of reasons.
Television series may need to be repeated, they must have characterizing 
``texture'' in each program's content, and support is needed for 
the distributed writing and production teams.   
For all these reasons the television program scripts represent
(technically) ideal and (application-wise) important 
exemplars for us.  As we will see below, we make full use of whatever
structure is available to us in such data.  

\subsection{Euclidean Embedding and Mapping of Context}
\label{sect13}

In our use of free text, a mapping into a 
Euclidean space gives us the capability to determine distance in a 
visual and easily interpretable way.  
In Correspondence Analysis (Murtagh 2005), the texts 
we are using -- e.g.\ the scenes -- 
provide the rows, and the set of terms used comprise the 
column set.  In the output, Euclidean, factor coordinate space, each text is 
located as a weighted average of the set of terms; and each term is located
as a weighted average of the set of texts.  (This simultaneous display is 
sometimes termed a biplot.)  So texts and terms are both 
mapped into the same, output coordinate space.  This can be of use in 
understanding a text through its closest terms, or vice versa.  Hence 
it provides semantic context for both the set of texts and the set of 
terms.  

A summary of properties of Correspondence Analysis is provided in the 
Appendix.  

A commonly used starting point for studying a set of texts, e.g.\ 
filmscript scenes, 
is to characterize each
text with numbers of terms appearing in the text, for a set of terms.

The $\chi^2$ distance
is an appropriate distance for use with such data
(Benz\'ecri 1979; Murtagh 2005).  The $\chi^2$ distance
is a weighted Euclidean distance.  
Consider texts $i$ and $i'$ crossed by words $j$.  Let $k_{ij}$ be the
number of
occurrences of word $j$ in text $i$.  Then, omitting a constant,
the $\chi^2$ distance between texts $i$ and $i'$ is given by
$ \sum_j 1/k_j ( k_{ij}/k_i - k_{i'j}/k_{i'} )^2$.  The weighting term 
here is
$1/k_j$.  The weighted Euclidean distance is between the {\em profile}
of text $i$, viz.\ $k_{ij}/k_i$ for all $j$, and the analogous
{\em profile} of text $i'$.  Our discussion is to within a constant because
we actually work on {\em frequencies} defined from the numbers of occurrences.
Define $f_{ij} = k_{ij} / k $ where $k = \sum_i \sum_j k_{ij}$.  Then the 
profile of scene $i$ is the set of values $f_{ij}$ for all $j$.  Similarly
the profile of word $j$ is the set of values $f_{ij}$ for all $i$.  We say 
that $f_i = \sum_j f_{ij}$ is the {\em mass} of scene $i$; and $f_j =
\sum_i f_{ij}$ is the mass of word 
$j$.  

Correspondence Analysis allows us to project the space of texts (we could
equally well explore the terms in the {\em same} projected space) into
a Euclidean space.  In doing so, Correspondence Analysis transforms 
(by a linear transform determined from a singular value decomposition) each 
pairwise $\chi^2$ distance into the
corresponding pairwise Euclidean distance. 

The Euclidean embedding is good for visualization, given that it is 
an intuitive one.  It is good for static context, including 
processing of a historical or otherwise sequenced set of information items.  
All inter-relationships are at once taken into consideration.  
Furthermore all such inter-relationships together provide context, 
relativities, and hence meaning.  In this sense therefore, 
Correspondence Analysis is an ideal platform for analysis of semantics
(Murtagh, 2009).  

\subsection{Television Filmscripts Used}

Production for television is often carried out by multiple teams.  
Notwithstanding this, there has to be a relatively very formulaic 
approach adopted to story and personalities.  Our goal is to see 
where and how tag clouds can express well the content of filmscript
data.  How can tag clouds provide us with a summary of the story 
and perhaps be displayed in conjunction with other forms of the 
story such as the movie itself?  We seek a better
summary than just pure word counts.  After all,
there is increasingly fast moving convergence in the sector, with 
successful television series being ported to a games environment
(this is the case for CSI); social network content being taken 
to a television environment (e.g.\ Sofia's Diary, www.bebo.com/sofiasdiary);
and so on.  

In this work we took three CSI (Crime Scene Investigation, 
Las Vegas -- Grissom, Sara,
Catherine  et al.) television scripts from series 1:

\begin{itemize}
\item 1X01, Pilot, original air date
on CBS Oct. 6, 2000.  Written by Anthony E. Zuiker, directed by Danny
Cannon.
\item 1X02, Cool Change, original air date on CBS, Oct.
13, 2000.  Written by Anthony E. Zuiker, directed by Michael Watkins.
\item 1X03, Crate 'N Burial,
original air date on CBS, Oct. 20, 2000.  Written by Ann Donahue, directed
by Danny Cannon.
\end{itemize}

Note the  differences between
writers and directors in most cases.  We will refer to these scripts as
CSI 101, CSI 102 and CSI 103.   All film scripts were obtained
from TWIZ TV (Free TV Scripts \& Movie Screenplays Archives),
http://twiztv.com

From series 3, we took another three scripts.

\begin{itemize}
\item 3X21, Forever, original air
date on CBS, May 1, 2003.  Written by Sara Goldfinger, directed by David
Grossman.
\item 3X22, Play With Fire, original air date on CBS,
May 8, 2003.  Written by Naren Shankar and Andrew Lipsitz, directed by
Kenneth Fink.
\item 3X23, Inside The Box, original air date on CBS, May 15,
2003.  Written by Carol Mendelsohn and Anthony E. Zuiker, directed by Danny
Cannon.
\end{itemize}

We will refer to these as CSI 321, CSI 322 and CSI 323.

An example of a very short scene, scene 25 from CSI 101, follows.

\begin{footnotesize}
\begin{verbatim}

[INT. CSI - EVIDENCE ROOM -- NIGHT]

(WARRICK opens the evidence package and takes out the shoe.)

(He sits down and examines the shoe.  After several dissolves, WARRICK opens the
lip of the shoe and looks inside.  He finds something.)

WARRICK BROWN:  Well, I'll be damned.

(He tips the shoe over and a piece of toe nail falls out onto the table.  He
picks it up.)

WARRICK BROWN:  Tripped over a rattle, my ass.

\end{verbatim}
\end{footnotesize}

We see here scene metadata, characters, dialog, and action information,
all of which we use.
Frontpiece, preliminary or preceding storyline information,
and credits were ignored by us.  
The number of scenes in each movie, and the number of
unique, 2-characters or more, words used in the movie, are listed in
Table \ref{tab0}.
All punctuation was ignored.  All upper case was
converted to lower case.  There was no pruning of stopwords (e.g., 
``the'', ``and'', etc.).  
In CSI 101 the top words and their frequencies of occurrence were:

\bigskip

\noindent the 443; to 239; grissom 195; you 176; and 166; gil 114;
catherine 105; of 89; he 85; nick 80; in 79; on 79; it 78; at 76;
ted 66; sara 65; warrick 65; ...

\bigskip

\begin{table}
\begin{center}
\begin{tabular}{ccc}  \hline
Script     &  No.\ scenes  &   No.\ words  \\ \hline
CSI 101    &     50        &    1679        \\
CSI 102    &     37        &    1343        \\
CSI 103    &     38        &    1413        \\
CSI 321    &     39        &    1584        \\
CSI 322    &     40        &    1579        \\
CSI 323    &     49        &    1445        \\ \hline
\end{tabular}
\end{center}
\caption{Numbers of scenes in the plot, and numbers of unique (2-letter
or more) words.}
\label{tab0}
\end{table}

The scenes constitute very natural segments of the filmscript.  
We determined frequencies of occurrence by 
scene of all words (subject only to what has been mentioned above
regarding case, punctuation, word size, etc.). 

\section{Capturing and Displaying Text Sequence Semantics through 
Spatial Embedding}
\label{sect2}

\subsection{Case Study: ``Crime Scene Investigation'' CSI 101 Pilot}
\label{sect23}

As an exemplary data set, we take a transcript
 from the widely-aired 
CSI, ``Crime Scene Investigation'', television series.  
The Pilot, 101, is used to begin with. 
In the Pilot, there are 50 scenes, with word counts ranging from 146 
words to 676 words.  In all there are 9934 words.  There are 1679 unique 
words, greater than 1 letter in length, with lower case replacing 
upper case, and with punctuation ignored.  We will use this 1679 unique 
word set.

The frequency of occurrence data crossing the 50 scenes and 1679 words is 
mapped, using Correspondence Analysis, into a space of intrinsic 
dimensionality 49: if $n, m$ are respectively the numbers of rows or 
scenes, and columns or words, then the inherent dimensionality is 
min($n-1, m-1$); the reason why 1 is subtracted from both is that the 
cloud of scenes and the cloud of words are both centered, giving a linear
dependence.  The origin is the average, expressing the hypothetical 
scene, or the hypothetical word, carrying no information.  

\begin{figure}
\includegraphics[width=15cm]{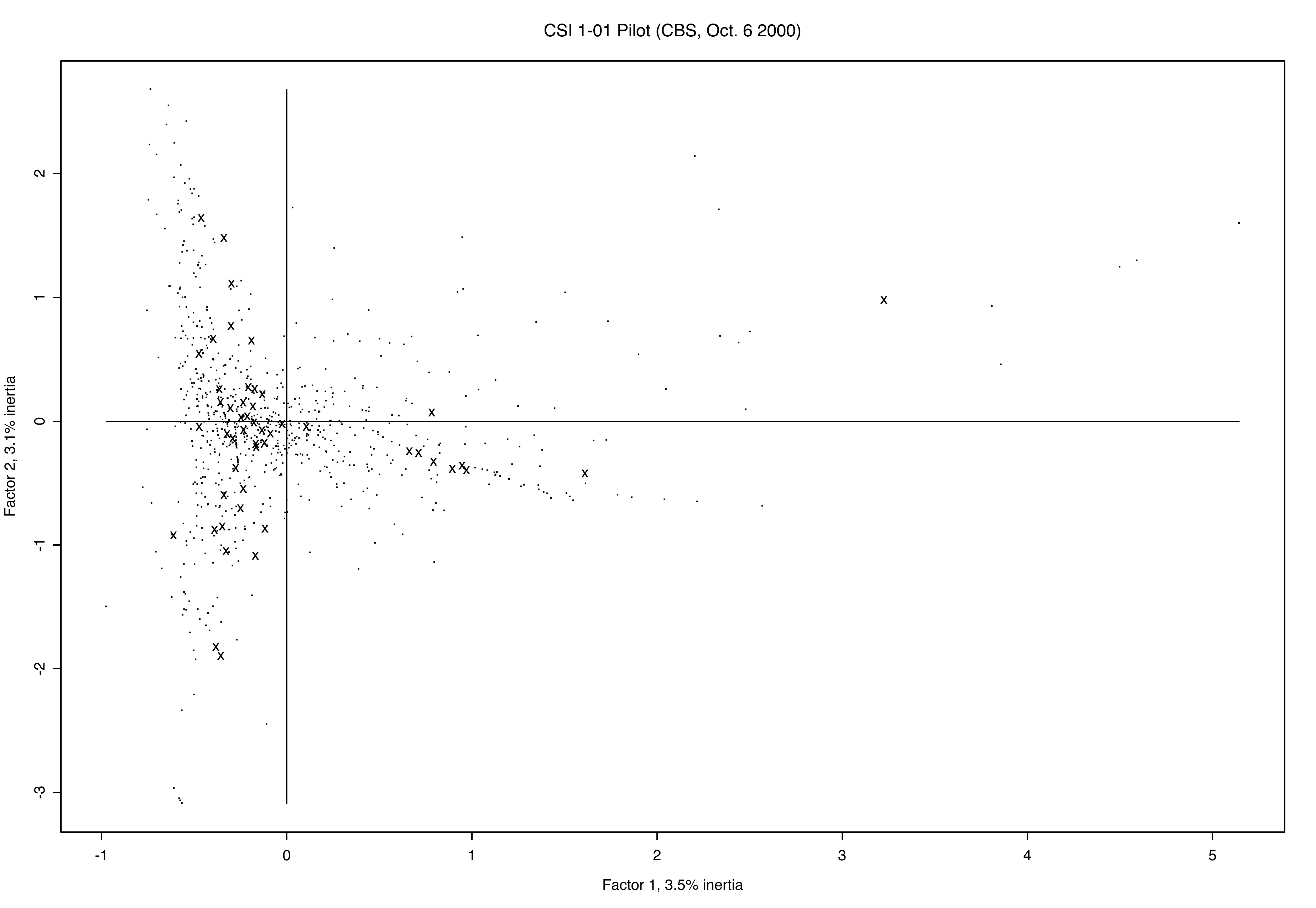}
\caption{Correspondence Analysis principal factor plane of projections 
of 50 scenes (each represented with an x), 
and 1679 characterizing words (each represented with a dot).  In this planar
view of the two clouds, the cloud of scenes and the cloud of words, we 
eschew labels for clarity.}
\label{csi101ca}
\end{figure}

In Figure \ref{csi101ca}, the scenes and words are located in the same
embedding.  The figure is interpreted in a visually natural, Euclidean, way, 
which is not the same as when we are presented with  
a frequency of occurrence data
array.  Defined on the basis of the frequency of occurrence array, 
we have the $\chi^2$ distance between scenes and/or
words.  The output display in Figure \ref{csi101ca} is a best 
planar view of a space endowed with the Euclidean metric.  Both scenes and 
words have a ``built-in'' normalization (defined above when we discussed
the $\chi^2$ metric).  

One important fact to 
keep in mind is that this is a best planar view of what is, in reality, a 
49-dimensional space.  The quality of the approximation involved in this 
is seen in the percentage inertia explained by these factors.  Inertia 
explained by a factor $r$ is the sum over all scenes of: mass 
times the projection squared on the axis.
Quite
typically for Correspondence Analysis, the extent of approximation is 
low in percentage terms.  
This is because less important factors or axes are ``explained 
by'', or determined by, isolated, very particular, words (which thereby
also determine the information content of particular scenes).  

The relationship between scenes and words in Figure \ref{csi101ca} is 
ultimately given by the {\em dual space} relationships (Murtagh 2005):
each scene is located at the center of gravity of all words; and each
word is located at the center of gravity of all scenes.  As noted
above, this establishes a semantic property for the location of any 
scene -- since the scene's location is determined by the word 
set.  Similarly each word's location is determined by the scene
set, hence establishing its semantics.  

In practice, Figure \ref{csi101ca} presents a very useful view of relationships
in our scenes $\times$ words data. We can look for polarities in the 
data; or anomalous scenes or words; or clusters or other configurations 
of scenes with reference to words or vice versa. 
But it is an approximation to the 
full dimensional reality.  Therefore for some purposes, as in the case of the
next section, we prefer to use the full dimensionality
Euclidean representation furnished by the factor space.  In such a case there
is no low dimensional projection involved, and no loss of information.  

\subsection{Selecting the Most Pertinent Terms}
\label{pertinent}

Presenting a result with around 1700 terms (cf.\ Figure \ref{csi101ca}) 
does not lend itself to 
convenient display.  We ask therefore what the most useful -- perhaps the 
most discriminating terms -- are.  In Correspondence Analysis both texts 
and their characterizing terms are projected into the same factor space.
So, from the factor coordinates, we can easily find the closest term(s) 
to a given text.  We do this for each of the 50 scenes and find -- in 
the full dimensionality factor space, and so with no approximation 
involved -- the following for the 50 scenes in succession in program 
CSI 101: 

\bigskip

\noindent
``royce''     ``soon''      ``coughs''    ``tape''      ``building''  ``makes''     ``gasps''     ``shift''     ``sign''      ``forced''   
``rushes''    ``city''      ``feet''      ``body''      ``hotel''     ``ah''        ``trying''    ``or''        ``business''  ``shoes''    
 ``screaming'' ``swab''      ``gun''       ``were''      ``rattle''    ``print''     ``really''    ``brass''     ``remember''  ``judge''    
 ``any''       ``latex''     ``skin''      ``both''      ``herself''   ``believe''   ``hospital''  ``dress''     ``finger''    ``minute''   
 ``deep''      ``statement'' ``minutes''   ``shh''       ``match''     ``second''    ``watching''  ``enters''    ``ring''      ``full''     

\bigskip
 
Among these terms, each individually characterizing a scene, in succession,
we note the following.  ``Royce'' is a personal name.  Terms like 
``Ah'' and ``Shh'' are present, and reflective of the scenes.  We 
experimented with other alternatives, such as selecting the more important
scenes -- which, from our procedure, leads also to the more important 
words -- on the basis of totalled high frequency of occurrence.  This did
not lead to a more attractive word set (e.g.\ nouns or verbs).  
Finally, we decided to limit ourselves to just one nearest
neighbor word for each scene on the grounds of facilitating interpretation.
However nothing restricts the consideration of, for example,
 3 or 4 nearest neighbors.  

In the following, 
in all cases, we use the first nearest neighbor word of 
each of the scenes.  We also discuss the use of a restricted set of words
in section \ref{sect32}.

\section{Application to Sematics-Based Tag Cloud Visualization}
\label{sect4}

\subsection{Television Transcripts}
\label{sect42}

\begin{figure}
\includegraphics[width=14cm]{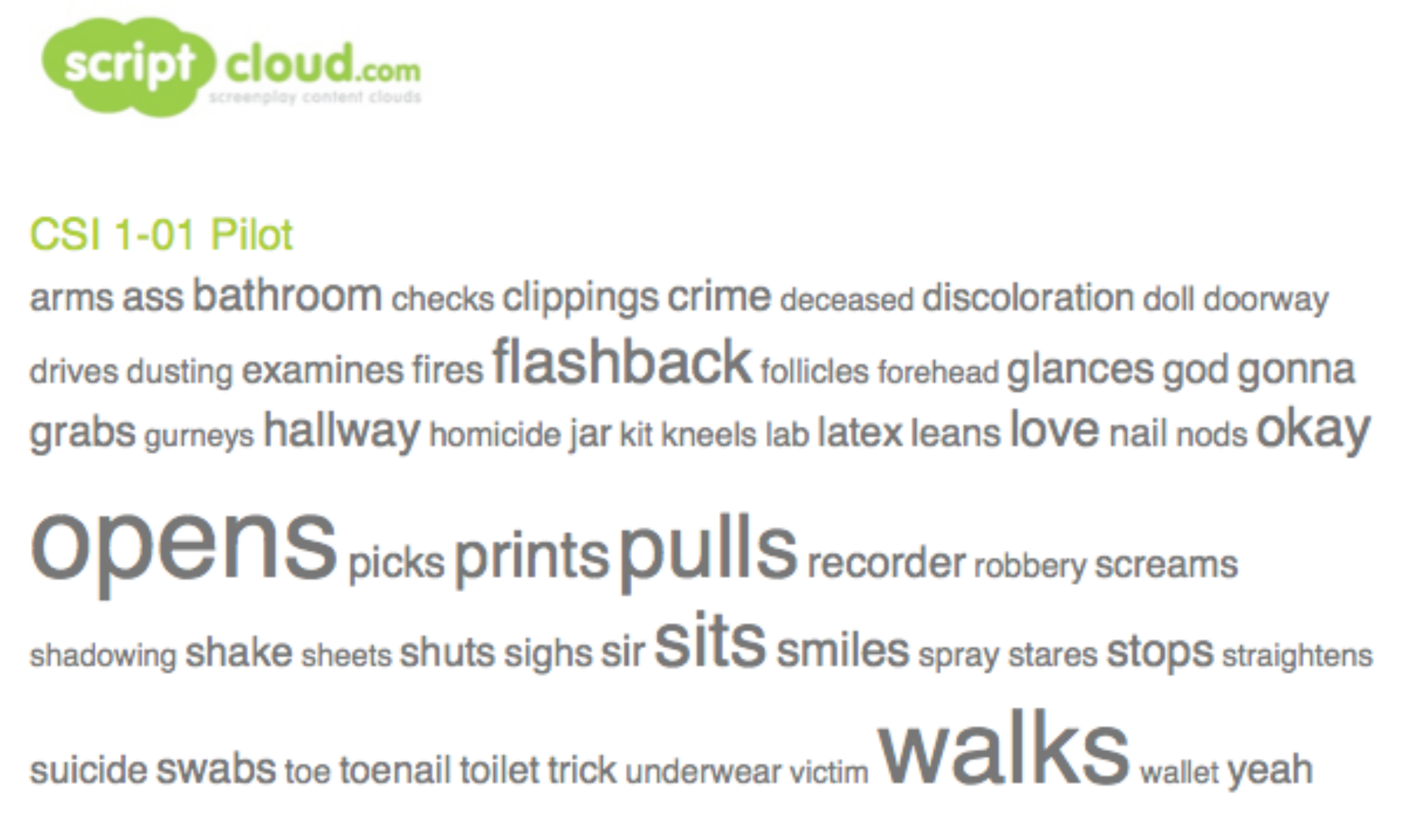}
\caption{A scriptcloud, showing 64 tags,
 based on frequent words retained following 
application of a stoplist.  Produced by an earlier version of 
Contentcloud, www.contentclouds.com,
using television program CSI 101.}
\label{scriptcloud}
\end{figure}

\begin{figure}
\includegraphics[width=14cm]{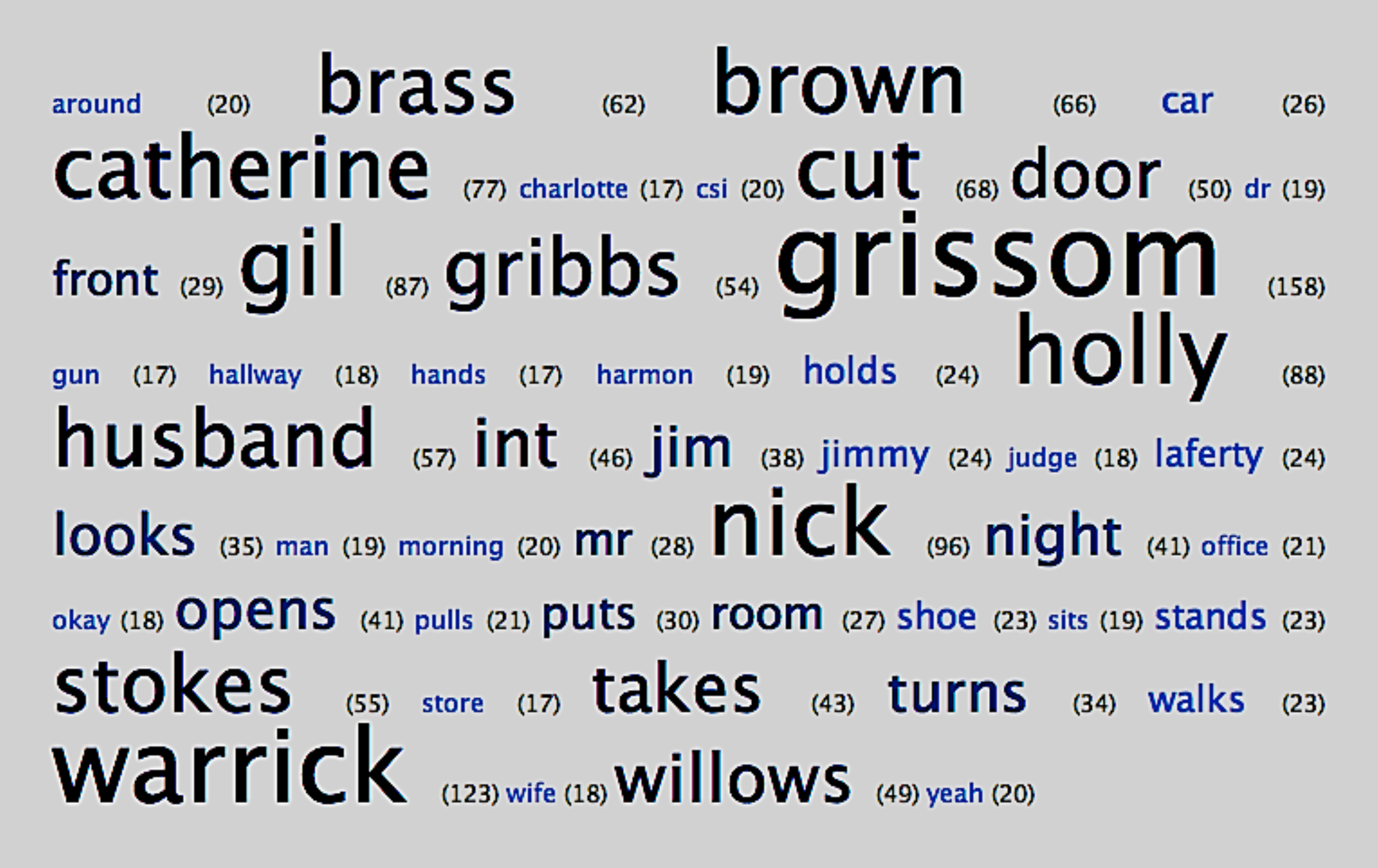}
\caption{A tag cloud, showing 50 tags, 
with stemming applied and very frequent words ignored.  Word frequencies
are also shown.  Produced
by TagCrowd, www.tagcrowd.com, using television program CSI 101.}
\label{tagcrowd}
\end{figure}

\begin{figure}
\includegraphics[width=14cm]{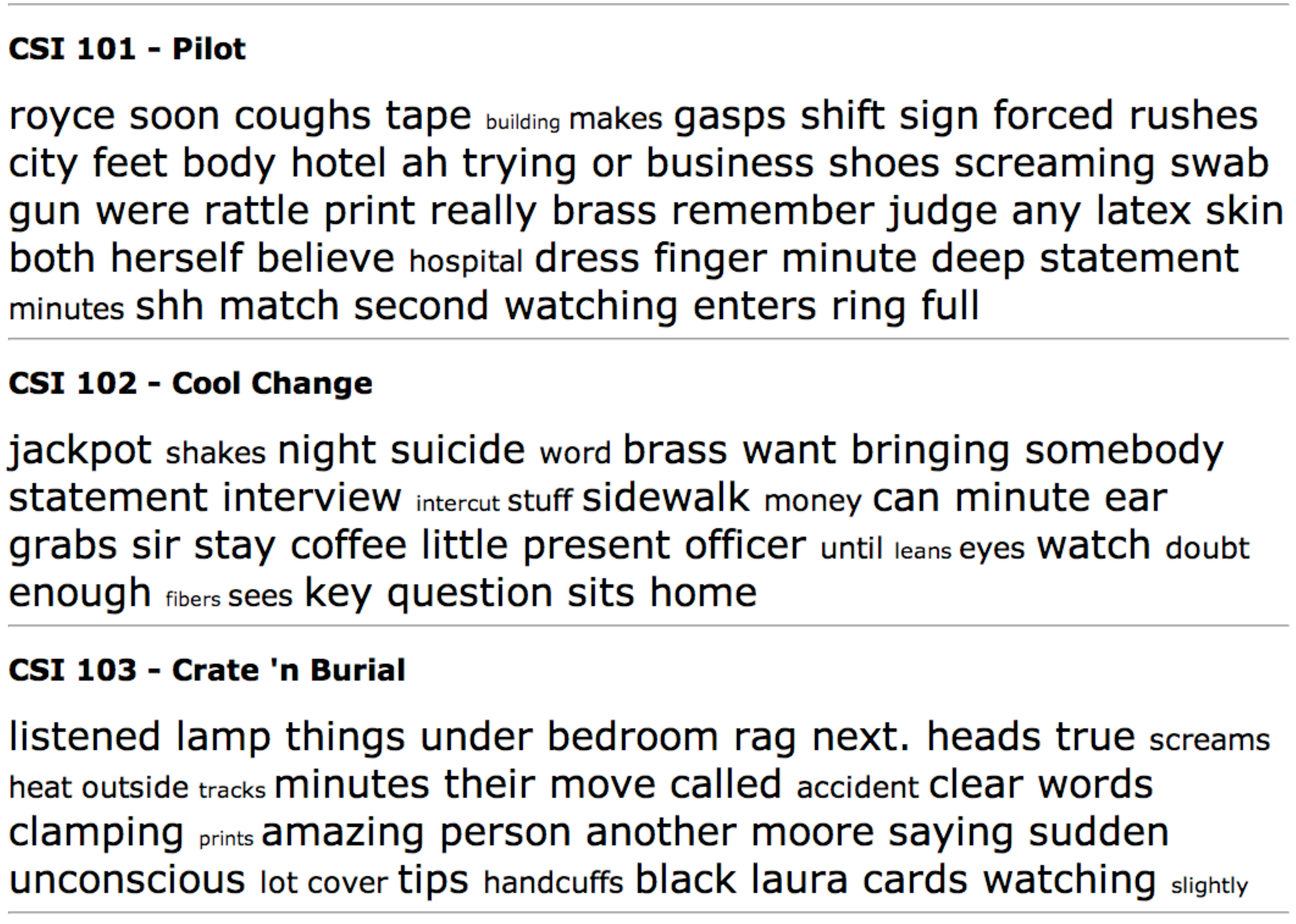}
\caption{Our tag cloud, where each word is a best characterization of a
scene.  The sequence of words corresponds to the sequences of scenes
(so the number of words displayed equals the number of scenes).
The word font size is proportional to semantic goodness of fit.
(See text for discussion.) 
Shown are, from top to bottom, the tag clouds for 
three television program scripts, CSI 101, CSI 102 and CSI 103.}
\label{mytagcloud}
\end{figure}

Let us now turn our attention to tag clouds.  To start with, we use
other implementations of 
tag clouds as an alternative approach to visualization, for comparison 
purposes.  Then we explore
how we can use the Correspondence Analysis opening up the 
filmscript semantics for us
in order to enhance the tag clouds.  As we will 
show tag clouds can express additional information about our filmscript data
while still maintaining the very clear display format.  

The tag clouds of  Figures \ref{scriptcloud} and \ref{tagcrowd}
use frequency information related
to word occurrence, and order words alphabetically.  There is nothing 
in such tag clouds that takes into account 
the sequential order of the original text.  The number of words is set by 
the user.  
   
Our tag cloud in Figure \ref{mytagcloud} orders words by scene, where the 
words are the best characterizations of the scenes,
in the sense described in section \ref{pertinent}.  Thus the most important 
structuring of the original text, viz.\ the sequence of scenes, is 
respected by this output display.  The number of words is the same as 
the number of scenes.  

In Figure \ref{mytagcloud} we define word size based on goodness of 
fit to the scene.  Since we have both words and scenes in the same
Euclidean embedding, goodness of fit to a scene is measured simply by 
Euclidean distance between word and scene.  To better scale the font 
sizes, and hence just for display purposes, we use a monotonic function 
of Euclidean distance, viz.\ log base 10 of the squared Euclidean distance.

Figure \ref{mytagcloud} 
also displays programs two and three of the first series 
of CSI.  The first of these programs has 50 scenes and 1679 
unique words; the second program has 37 scenes and 1343 words; and the
third program has 38 scenes and 1413 words.  

In the top panel of Figure \ref{mytagcloud} we note that most of the terms
are roughly equally close to respective scenes
(i.e.\ in the same band of closeness, since 
we allowed for six bands defining six font sizes) but not the term
``building'' in scene 5, nor ``minutes'' in scene 43.  So, as good 
representatives of the corresponding scenes, these terms are in the 
nature of ``handle with caution'' matches.  They are however, by 
definition, the semantically closest terms.  

To summarize, we 
started with terms that most closely expressed and characterized the
scenes, and indeed that discriminated, by a nearest neighbor definition,
between the scenes.  Representativity of a scene by a word was expressed
by semantic proximity.  
All aspects of this are based on a Euclidean embedding, which also 
directly addresses the normalization of the original statistical 
(i.e.\ frequency of occurrence) data, both in relation to words and in 
relation to scenes.  Furthermore all aspects of what we have done is 
automated and does not require any user setting of thresholds or manual 
selections.  

To compare these results with other program (or episode) 
scripts, we can avail of 
other data available at TWIZ (2007). (Seasons 1 through 4 of CSI have each 
23 programs; season 5 has 25 programs; seasons 6 and 7 have 24 programs;
and season 8 has 17 programs.)

We use the following (Table \ref{tab0}):

\begin{itemize}

\item CSI 321, 
third series, 21st program, ``Forever'', originally aired on CBS on 1 May 2003,
39 scenes.   (In all, 1584 unique words were used.) 

\item 
CSI 322, third series, 22nd program, ``Play with Fire'', 
  originally aired on CBS on 8 May 2003, 40 scenes.  (In all, 1579 words
were used.) 

\item
CSI 323, third series, 
23rd and last program of the series, ``Inside the Box'', 
  originally aired on CBS on 15 May 2003, 49 scenes. (In all, 1445 words 
were used.)  
\end{itemize}

\begin{figure}
\includegraphics[width=14cm]{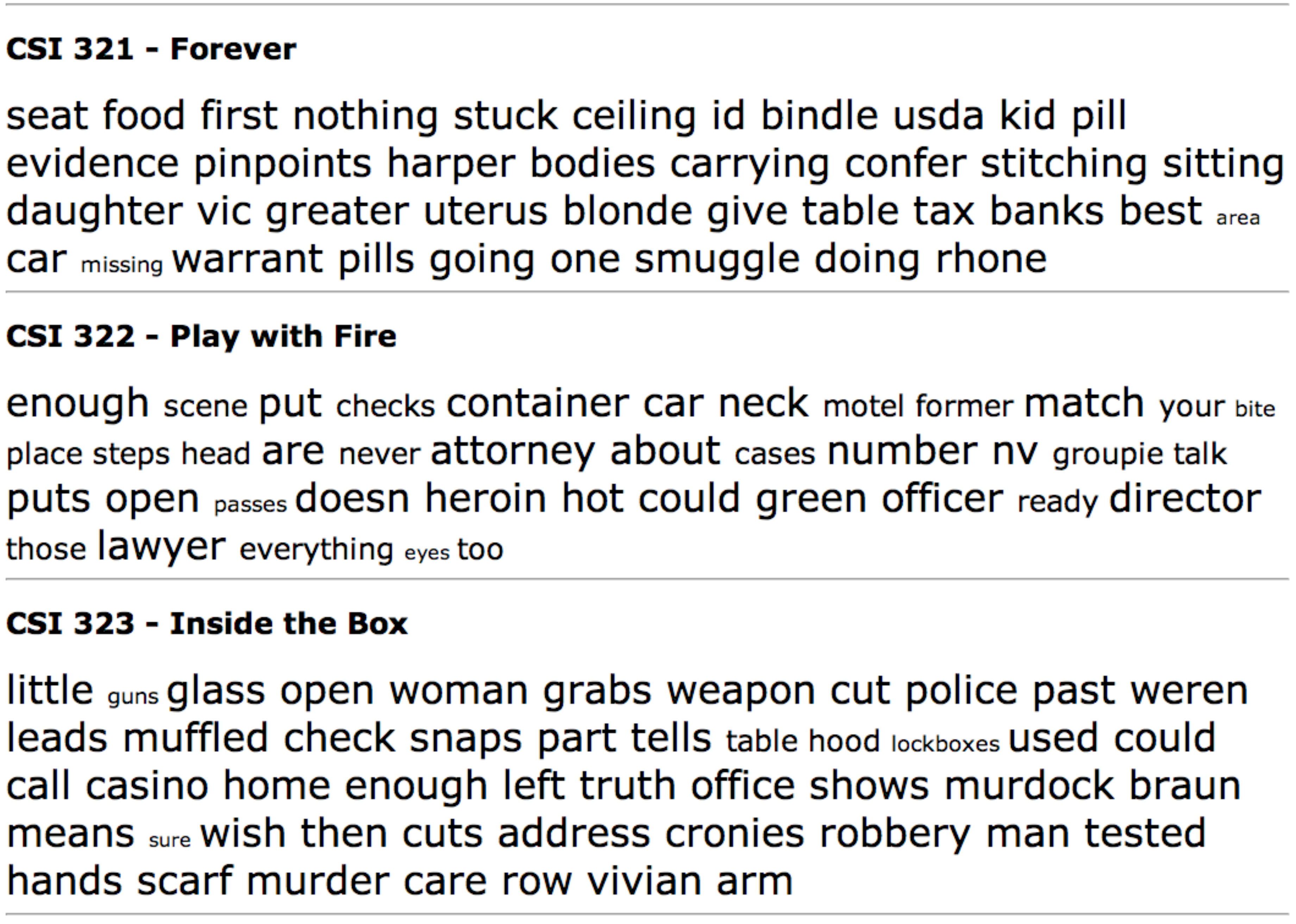}
\caption{Tag clouds for programs (episodes) CSI 321, CSI 322 and 
CSI 323 -- the final ones
-- in CSI Series 3.  See Figure \ref{mytagcloud} for other details.}
\label{mytagcloud2}
\end{figure}

Figure \ref{mytagcloud2} shows the resulting tag clouds.  
Some remarks on this figure follow.
``NV'' (CSI 322) is part of a much used identifier of personages in a Nevada 
Correctional Facility in scene 22.  Also in CSI 322, our punctuation handling
has left ``doesn'' and stripped the remainder.  
We can of course base scene characterization 
on a selected set of words and not all words.  In the section to follow, 
section \ref{sect32}, we do just that.  

In our tag clouds, as explored in Figures \ref{mytagcloud} and 
\ref{mytagcloud2}, input terms are chosen as the nearest neighbor 
word for each scene. 
In these visualizations we have the following simultaneous properties:

\begin{itemize}

\item  Selected words, that most appropriately characterize a scene.
\item The selected words accompanying the sequence of scenes, in scene order.
\item Quality of the fit between characterizing word and scene.
\item All pairwise relationships of words and scenes taken into consideration.
\end{itemize}

Properties such as these, seen in the 
display, can be used to distinguish one episode (program) from another.  

One of the most appealing aspects of our approach is 
that all phases of the processing 
are automated, and there are no user-set parameters or other user 
intervention required.

We checked that the pertinent words found to characterize each scene were 
unique.  This was always the case.  If the same word were found to be closest
to two scenes we could of course use any such multiple occurrences of words.  

The motivation for our tag clouds is to go beyond frequency of 
occurrence statistics and instead visualize at once multiple facets 
of the filmscripts.  

\subsection{Tracking of Characters}
\label{sect32}

In the sequence of scenes balance must be maintained as well as 
tempo-related contrast.  In such areas as contrasts between interior and 
exterior scenes, day and night, and the presence or absence of 
principal and secondary characters, the filmscript must reflect 
vital aids and hints to the viewer, provoking both continuity of 
understanding by the viewer and discontinuity to trigger hightened 
attention.  

We will look at the principal characters in the CSI scripts and 
television series programs: Gil Grissom, Warrick Brown, 
Nick Stokes, Catherine Willows, Jim Brass, and Sara Sidle.  We will 
refer to them by the first or family name mainly used: Grissom, Warrick,
Nick, Catherine, Brass and Sara.  

\begin{figure}
\includegraphics[width=14cm]{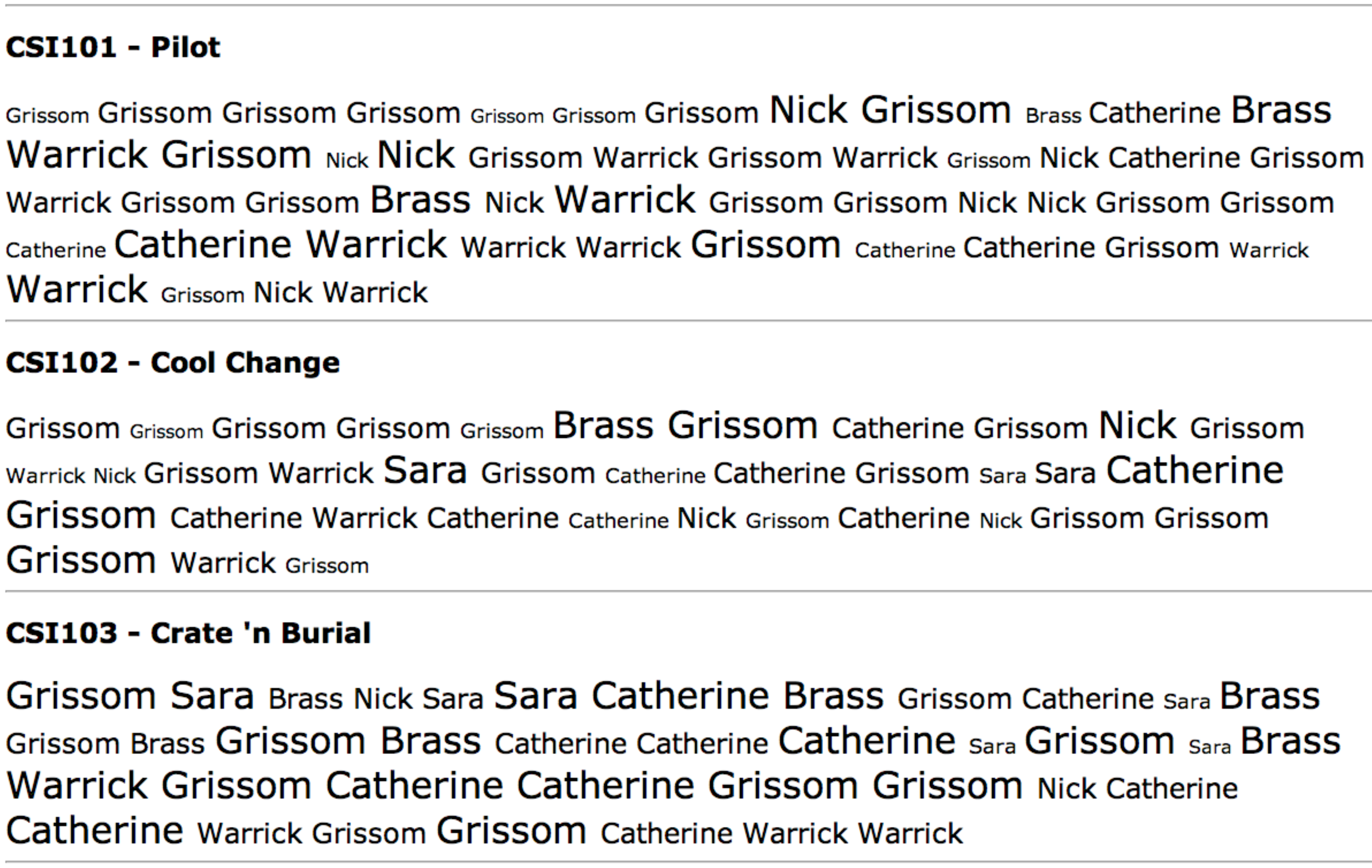}
\caption{In CSI 101, CSI 102 and CSI 103, there were respectively 
50, 37 and 38 scenes.  We show the most important character, among 
the six principal characters, for each scene in succession.  The size
used in the display, expressing relative importance for the scene,
 is defined via proximity between scene and character name, 
as explained in the text.}
\label{mytagcloud-nameslog}
\end{figure}

\begin{figure}
\includegraphics[width=14cm]{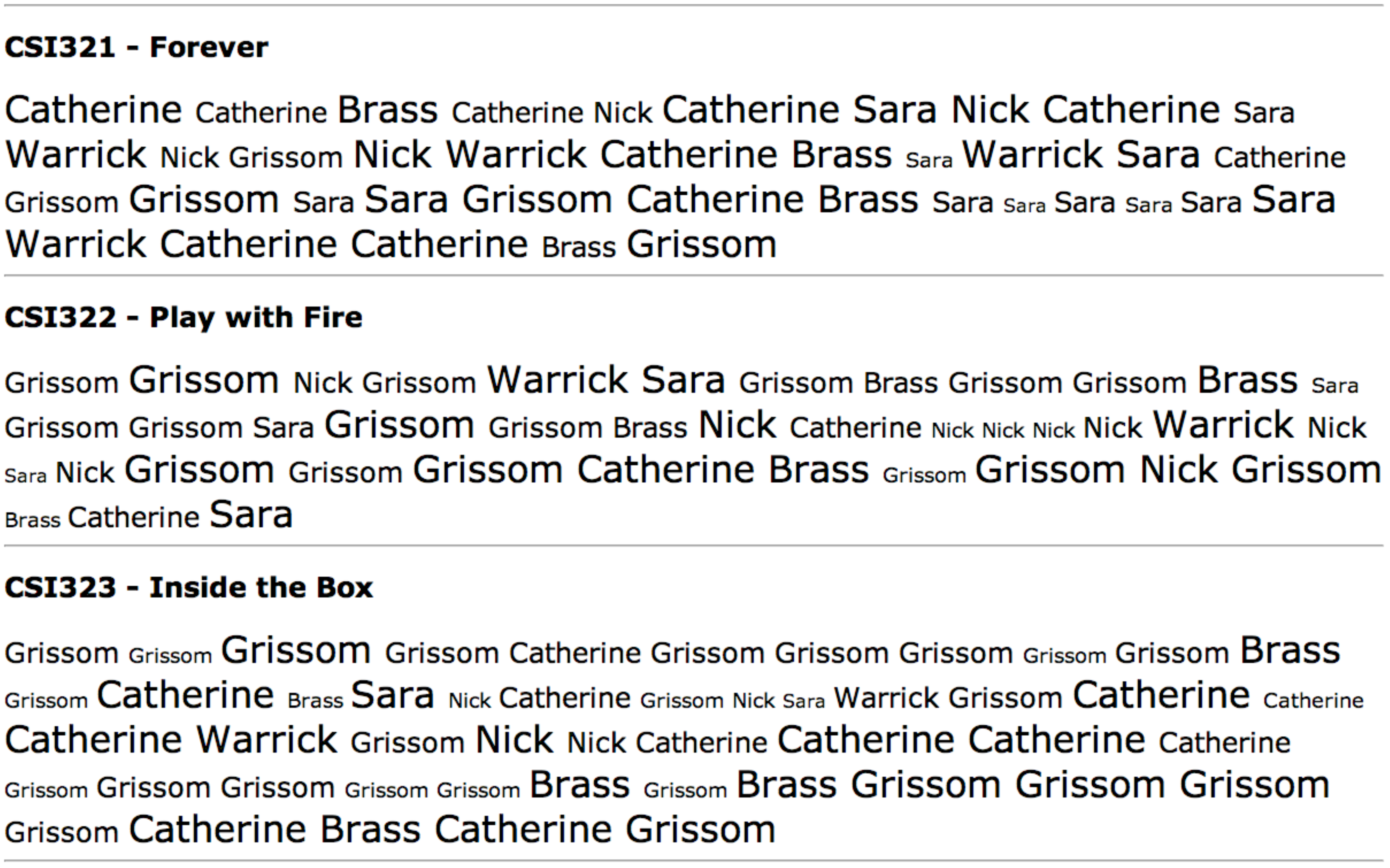}
\caption{In CSI 321, CSI 322 and CSI 323, there were respectively 
39, 40 and 49 scenes.  We show the most important character, among 
the six principal characters, for each scene in succession.  The size
used in the display, expressing relative importance for the scene,
 is defined as explained in the text.}
\label{mytagcloud-nameslog2}
\end{figure}

The Correspondence Analysis allows us to easily seek the principal 
character who is closest to each scene.  In the plot of scenes crossed
by all words used in the filmscript, which naturally contains the 
character names, we look for proximity -- in the full dimensional 
Euclidean, factor space, so no approximation is involved -- between 
the character and the scenes.  The relative importance is expressed 
by size in Figures \ref{mytagcloud-nameslog} and \ref{mytagcloud-nameslog2}.
This relative importance is a scaled version of the log (base 10) 
of the squared Euclidean distance.  (Using the distance or 
squared distance, and taking the log, there is clearly no effect
on monotonicity of 
proximity.  We take the log for improved visual appearance.)  

We can see at a glance how Grissom pervades these films; whether 
characters reappear as the most crucial players implying intertwining 
of different actors; how the central roles of male and female characters
alternate; and so on.  We could of course collect statistics of 
appearance, and present such results as histograms or pie charts,
or a time series.  However the motivation for our tag clouds is to have
a range of properties of the filmscript presented simultaneously.   

\section{Conclusions}

Our objective in this work has been to visualize the semantics
of filmscript, in a novel way, by relying on quite a widely
used approach to visualization, namely tagclouds.  We have 
related this visualization to a comprehensive framework which 
allows us to go further when and where needed, such as to 
find further neighborhood relationships between terms.
Filmscript, our input data, offers us a useful testbed to 
prototype our new algorithms.  Analysis and interpretation of 
filmscript has direct and, in fact, far-reaching implications
for the evolving production process for filmscript, and the 
changing world of film, television, games, and all online media.  
Beyond filmscript, our work has possible relevance for  any 
other partially structured, time-ordered, sequence of text segments.  

With reference to Chafe (1979) we used the sequence of text segments 
representing scenes in the filmscript.
Sequence and semantic proximity are presented simultaneously in our
tag cloud display.  

Computationally, all processing is of linear time in the scenes,
or their associated words.  The eigen-reduction at the core of the 
Correspondence Analysis has a cubic computation time in either 
the scene set or the word set: of course we choose to carry out this
computationally heavy processing on the smaller of these two sets, 
viz.\ the set of scenes.   Once this is done we easily pass between 
the dual spaces of scenes and words (see Appendix).

Our approach is automated, without recourse to user parameters, or user
choice of chained tasks.  A framework enveloping input data and 
delivered (potentially interactive and responsive) display is provided.

Our comparisons in this work 
have included (i) basic low dimensional mapping
and (ii) two other tag cloud visualizations.  We build 
on these approaches in order to have a better visualization.  The 
broad context for our work is that of support for film production for 
television. As noted, this is an area of great commercial importance,
not least in view of the convergence between Internet, television,
cinema, games, print media, mobile communications and other 
display devices, and so on.  

As noted too, we use filmscripts which are the points of departure 
for a later film, or for the film's accompanying
metadata.  The original semantics of
the television programs are used with further processing 
provided through the Correspondence 
Analysis mapping. 

We find our approach to be particularly useful when words are selected, 
such as when our focus is on the dramatic characters (personalities).  

\section*{Appendix: Correspondence Analysis}

\subsection*{Analysis Chain}

Correspondence Analysis 
provides what could be characterized as a data analysis platform
providing access to the semantics of information expressed by the data.
The way it does this is by viewing each observation or row vector as the
average of all attributes that are related to it; and by viewing each
attribute or column vector as the average of all observations that are
related to it.

The analysis chain is as follows:

\begin{enumerate}
\item The starting point is a matrix that cross-tabulates the dependencies,
e.g.\ frequencies of joint occurrence, of an observations crossed by attributes
matrix.
\item By endowing the cross-tabulation matrix with the $\chi^2$ metric
on both observation set (rows) and attribute set (columns), we can map
observations and attributes into the same space, endowed with the Euclidean
metric.
\item Interpretation is through projections of observations, attributes
or clusters onto factors.  The factors are ordered by decreasing importance.
\end{enumerate}

There are various aspects of Correspondence Analysis which follow on
from this, such as Multiple Correspondence Analysis, different ways that
one can encode input data, inducing of a hierarchical clustering from 
the (Euclidean) factor space, and mutual description of clusters in terms of
factors and vice versa.  See Murtagh (2005) and references therein
for further details.

We will use a very succinct and powerful tensor notation in the following,
introduced by Benz\'ecri (1979).  At key points we will indicate the
equivalent vector and matrix expressions.

\subsection*{Correspondence Analysis:  
Mapping $\chi^2$ Distances into Euclidean Distances}

The given contingency table (or numbers of occurrence)
data is denoted $k_{IJ} = 
\{ k_{IJ}(i,j) = k(i, j) ; i \in I, j \in J \}$.  $I$ is the set of
observation
indexes, and $J$ is the set of attribute indexes.  We have
$k(i) = \sum_{j \in J} k(i, j)$.  Analogously $k(j)$ is defined,
and $k = \sum_{i \in I, j \in J} k(i,j)$.  Next, $f_{IJ} = \{ f_{ij} 
= k(i,j)/k ; i \in I, j \in J\} \subset \R_{I \times J}$,
similarly $f_I$ is defined as  $\{f_i = k(i)/k ; i \in I, j \in J\} 
\subset \R_I$, and $f_J$ analogously.  What we have described here is
taking numbers of occurrences into relative frequencies.

The conditional distribution of $f_J$ knowing $i \in I$, also termed
the $j$th profile with coordinates indexed by the elements of $I$, is:

$$ f^i_J = \{ f^i_j = f_{ij}/f_i = (k_{ij}/k)/(k_i/k) ; f_i > 0 ; 
j \in J \}$$ and likewise for $f^j_I$.

\subsection*{Input: Cloud of Points Endowed with the Chi Squared Metric}

The cloud of points consists of the couples:
(multidimensional) profile coordinate and (scalar) mass.
We have $ N_J(I) = \{ ( f^i_J, f_i ) ; i  \in I \} \subset \R_J $, and
again similarly for $N_I(J)$.  Included in this expression is the fact
that the cloud of observations, $ N_J(I)$, is a subset of the real
space of dimensionality $| J |$ where $| . |$ denotes cardinality
of the attribute set indexed by $J$.

The overall inertia is as follows:
$$M^2(N_J(I)) = M^2(N_I(J)) = \| f_{IJ} - f_I f_J \|^2_{f_I f_J} $$
\begin{equation}
= \sum_{i \in I, j \in J} (f_{ij} - f_i f_j)^2 / f_i f_j
\label{eqnin}
\end{equation}
The term  $\| f_{IJ} - f_I f_J \|^2_{f_I f_J}$ is the $\chi^2$ metric
between the probability distribution $f_{IJ}$ and the product of marginal
distributions $f_I f_J$, with as center of the metric the product
$f_I f_J$.  Decomposing the moment of inertia of the cloud $N_J(I)$ -- or
of $N_I(J)$ since both analyses are inherently related -- furnishes the
principal axes of inertia, defined from a singular value decomposition.

\subsection*{Output: Cloud of Points Endowed with the Euclidean 
Metric in Factor Space}

The $\chi^2$ distance with center $f_J$ between observations $i$ and
$i'$ is written as follows in two different notations:

\begin{equation}
d(i,i') = \| f^i_J - f^{i'}_J \|^2_{f_J} = \sum_j \frac{1}{f_j}
\left( \frac{f_{ij}}{f_i} - \frac{f_{i'j}}{f_{i'}} \right)^2
\end{equation}

In the factor space this pairwise distance is identical.  The coordinate
system and the metric change.  For factors indexed by $\alpha$ and for
total dimensionality $N$ ($ N = \mbox{ min } \{ |I| - 1, |J| - 1 \}$;
the subtraction of 1 is since
the $\chi^2$ distance is centered and
hence there is a linear dependency which
reduces the inherent dimensionality by 1) we have the projection of
observation $i$ on the $\alpha$th factor, $F_\alpha$, given by
$F_\alpha(i)$:

\begin{equation}
d(i,i') = \sum_{\alpha = 1..N} \left( F_\alpha(i) - F_\alpha(i') \right)^2
\end{equation}

In Correspondence Analysis the factors are ordered by decreasing
moments of inertia.  The factors are closely related, mathematically,
in the decomposition of the overall cloud,
$N_J(I)$ and $N_I(J)$, inertias.  The eigenvalues associated with the
factors, identically in the space of observations indexed by set $I$,
and in the space of attributes indexed by set $J$, are given by the
eigenvalues associated with the decomposition of the inertia.  The
decomposition of the inertia is a
principal axis decomposition, which is arrived at through a singular
value decomposition.

\subsection*{Dual Spaces and Transition Formulas}

The projection of observation $i$ on the $\alpha$th 
factor is $F_\alpha(i)$.  Likewise, the projection of attribute 
$j$ on the $\alpha$th factor in its associated space is $G_\alpha(j)$.  
The following relationship holds.  

$$ F_\alpha(i) = \lambda^{-\frac{1}{2}}_\alpha \sum_{j \in J}                  
f^i_j G_\alpha(j) \mbox{  for  } \alpha = 1, 2, \dots N; i \in I$$

\begin{equation} 
G_\alpha(j) = \lambda^{-\frac{1}{2}}_\alpha \sum_{i \in I}                  
f^j_i F_\alpha(i) \mbox{  for  } \alpha = 1, 2, \dots N; j \in J
\end{equation}

Relation (4) gives us the {\em transition formulas}:  
The coordinate of
element $i \in I$ is the barycenter of the coordinates of the elements
$j \in J$, with associated masses of value given by the coordinates of
$f^i_j$ of the profile $f^i_J$.  This is all to within the
$\lambda^{-\frac{1}{2}}_\alpha$ constant.

In the
output display, the barycentric principle comes into play: this allows
us to simultaneously view and interpret observations and attributes.

\section*{References}
\begin{description}

\item[]
Beard K, Deese H and Pettigrew NR.
A framework for visualization and exploration of events, 
Information Visualization, 7, 133-151, 2008.

\item[] 
Benz\'ecri J-P.  {\em L'Analyse des Donn\'ees, Tome I Taxinomie,
Tome II Correspondances}, 2nd ed.\ Dunod, Paris,  1979.

\item[]
Chafe WL.  The flow of thought and the flow of language. In 
T. Giv\'on, Ed., {\em Syntax and Semantics: Discourse and Syntax}, 1979,
Vol. 12, Academic Press, 159--181 

\item[] Doyle LB. Semantic road maps for literature searches.
Journal of the ACM, 8, 553--578, 1961.

\item[]
Gladwell M. 
The formula: What if you build a machine to predict hit movies?
The New Yorker, October 16, 2006 \\
http://www.newyorker.com/archive/2006/10/16/061016fa\_fact6

\item[]
Hearst M.  Multi-paragraph segmentation of expository text, 
Annual Meeting of the ACL, Proc. 32nd Annual Meeting of Association for 
Computational Linguistics,  9--16, 1994.  

\item[]
Kaser O, Lemire D.  Tag-cloud drawing: algorithms for cloud
visualization.  Proc.\ Tagging and Metadata for Social Information
Organization, WWW 2007.  arxiv.org/pdf/cs/0703109v2

\item[]
Kohonen T, Kaski S, Laski K, Salj\"arvi J, Honkela J, Paatero V, 
Saarela A. Self-organization of a massive document collection. 
IEEE Transactions on Neural Networks, 11, 574--585, 2000.

\item[] McKie S. 
Scriptcloud: content clouds for screenplays.
Report, 2007.  See also: Screenplay content clouds, www.contentclouds.com

\item[] Merali Z.
Here's looking at you, kid. Software promises to identify blockbuster scripts,
Nature, 453, p.\ 708, 2008.  

\item[] Murtagh F.  {\em Correspondence Analysis and Data 
Coding with R and Java}, Chapman \& Hall/CRC, Boca Raton, 2005.

\item[] Murtagh F.  Visual user interfaces, interactive maps, 
references,
theses, \\
http://astro.u-strasbg.fr/$\sim$fmurtagh/inform, 2006.

\item[] Murtagh F. The Correspondence Analysis platform for 
uncovering deep structure in data and information, Sixth Boole 
Lecture. Computer Journal, forthcoming, 2009. 
Advance Access 9 Sept.\ 2008, doi:10.1093/comjnl/bxn045.

\item[] Murtagh F, Taskaya T, Contreras P, Mothe J, Englmeier K.
Interactive visual user interfaces: a survey, Artificial Intelligence 
Review, 19, 263--283, 2003. 

\item[]
Murtagh F, Ganz A and McKie S. 
The structure of narrative: the case of film scripts. Pattern 
Recognition, 42, 302--312, 2009.

\item[]
Olston C. and Chi, E.H. ScentTrail: integrating browsing and 
searching on the web. ACM Transactions on Computer-Human 
Interaction, 10, 177--197, 2003. 

\item[]
TWIZ TV, CSI Transcripts, Seasons 1 to 8, www.twiztv.com/scripts/csi 

\end{description}

\end{document}